# From Propagator to Oscillator: The Dual Role of Symmetric Differential Equations in Neural Systems

Kun Jiang            atan_j@qq.com

*School of Electrical Engineering*

*Chongqing University*

*Shazheng St.174, Shapingba, China*

**Abstract**

In our previous work, we proposed a novel neuron model based on symmetric differential equations and demonstrated its potential as an efficient signal propagator. Building upon that foundation, the present study delves deeper into the intrinsic dynamics and functional diversity of this model. By systematically exploring the parameter space and employing a range of mathematical analysis tools, we theoretically reveal the system's core property of functional duality. Specifically, the model exhibits two distinct trajectory behaviors: one is asymptotically stable, corresponding to a reliable signal propagator; the other is Lyapunov stable, characterized by sustained self-excited oscillations, functioning as a signal generator. To enable effective monitoring and prediction of system states during simulations, we introduce a novel intermediate-state metric termed on-road energy. Simulation results confirm that transitions between the two functional modes can be induced through parameter adjustments or modifications to the connection structure. Moreover, we show that oscillations can be effectively suppressed by introducing external signals. These findings draw a compelling parallel to the dual roles of biological neurons in both information transmission and rhythm generation, thereby establishing a solid theoretical basis and a clear functional roadmap for the broader application of this model in neuromorphic engineering.

1. Introduction

Artificial Neural Networks (ANNs), particularly deep learning systems, have achieved remarkable breakthroughs in recent years across a wide range of fields including image recognition, language understanding, control, and decision-making[1-3]. However, these successes largely rely on massive computational resources and high energy consumption, which significantly limits their deployment in edge computing and low-power embedded environments[4, 5]. In sharp contrast, biological neural systems process information with extraordinary efficiency and adaptability, operating in a massively parallel fashion at extremely low energy cost[6, 7]. This striking gap in energy efficiency has prompted researchers to reconsider the foundational design of artificial neural networks and explore models that exhibit biological plausibility[8, 9].

Biological plausibility refers to the extent to which a neural network model aligns with biological neural systems in terms of dynamics, hierarchical structure, and learning mechanisms[10]. Particularly at the dynamical level, biological neurons exhibit complex nonlinear behaviors, such as depolarization, threshold firing, self-excited oscillations, and post-inhibitory rebound, which cannot be captured by simple linear summation[11]. While the classical Hodgkin–Huxley (HH) model provides a detailed and accurate description of membrane potential dynamics[12, 13], its complexity and analytical intractability hinder its application in large-scale neural simulations. Consequently, a variety of simplified yet computationally tractable models have been proposed, such as the Leaky Integrate-and-Fire (LIF), FitzHugh–Nagumo (FHN), and Izhikevich models, aiming to balance biological realism with mathematical manageability.

Among these models, differential equations play a central role in describing the temporal evolution of neural states[14]. By carefully designing dynamical structures and coupling relationships among variables, differential equations naturally reproduce various neural phenomena, including spiking, synaptic integration, and rhythmic activity propagation across network layers. For instance, the FHN model simplifies the HH model into a two-dimensional dynamical system that captures the interaction between excitation and recovery variables. The Izhikevich model introduces nonlinear control terms that enable rich pattern generation even in low-dimensional systems. Modern computational neuroscience further extends this modeling paradigm by employing delay differential equations, stochastic differential equations, and fractional-order systems to mimic synaptic delays, stochastic firing, and other biologically realistic features.

Beyond signal response and transmission, many neurons in biological neural networks also exhibit rhythmic generation, autonomously producing oscillatory activity that is critical for a wide array of neural functions[15]. This capability is indispensable in several key systems:

Respiratory control neurons: Pacemaker neurons in the brainstem can generate rhythmic breathing patterns independently and maintain respiratory activity through spinal motor circuits.

Central Pattern Generators (CPGs): Found in insects, fish, and mammals, CPGs orchestrate rhythmic motor patterns such as walking, swimming, and chewing, even in the absence of sensory feedback[16].

Thalamic oscillatory neurons: These neurons generate low-frequency rhythms that interact with cortical networks and contribute to processes such as sleep and attention modulation.

Neurons in such systems often display strong self-sustained oscillations, typically exhibiting Lyapunov stability without asymptotic convergence, as their trajectories revolve around a stable limit cycle[17]. These neurons function as signal generators, producing endogenous rhythmic activity essential for initiating behavior, controlling movement, and coordinating rhythmic processes. As a result, emulating this signal-generating capability has become a critical challenge in fields such as neuromorphic computing, brain–machine interfaces, and intelligent control systems[18].

Although some existing models are capable of exhibiting oscillatory behavior under specific parameter settings, they often rely on ad hoc nonlinear activation functions, gating mechanisms, or intricate high-dimensional state embeddings[19, 20]. These approaches lack a unified, concise, and analytically tractable modeling framework. Consequently, a central open question in biologically inspired neural modeling is how to construct a single model that can serve both as a signal propagator and as a signal generator.

To address this challenge, we propose a novel approach based on the principle of symmetry, leading to the development of a symmetric differential equation that serves as the foundation for a scalable biologically inspired neural framework. In our prior studies [21-23] we have demonstrated that this framework functions effectively as a signal propagator, enabling stable information processing and successful training in large-scale networks. These results provide preliminary evidence for the model's mathematical soundness, scalability, and biological plausibility[24, 25].

Yet an important question remains: Is this model limited to signal propagation alone? Biology clearly indicates otherwise—many vital life functions, such as breathing and locomotion, depend on pacemaker neurons and central pattern generators capable of autonomously producing rhythmic outputs. This raises a critical inquiry: Can our proposed symmetric differential model also emulate the signal generator functionality observed in biological neurons?

In this work, we provide a definitive affirmative answer to this question, supported by both theoretical analysis and numerical simulations. We examine the core dynamical structure of the model and rigorously characterize its functional duality. The structure of the paper and its main contributions are as follows:

Chapter 2 introduces the symmetric differential equation and systematically explores its feasible parameter space, laying the mathematical foundation for subsequent stability analysis and functional transitions.

Chapter 3 presents in-depth theoretical analysis. Using Lyapunov's second method, Brouwer's fixed-point theorem, and homotopy arguments, we rigorously prove that system trajectories exhibit one of two behaviors: asymptotic stability (associated with signal propagation) or Lyapunov stability (associated with oscillatory signal generation)[26, 27]. These results demonstrate that the system's trajectories either converge to a unique fixed point or repeatedly return to its vicinity.

Chapter 4 validates these theoretical findings through numerical simulations. We illustrate how modifications to parameters or connection patterns can induce a switch between stable propagation and rhythmic generation. During this process, we introduce a novel metric—on-road energy—to dynamically monitor the system's evolution. Furthermore, we explore how external signal injection can be used to suppress oscillations and stabilize the system, showcasing the model's expressive power and controllability.

Chapter 5 concludes the paper and discusses potential applications of the model in neuromorphic computing.

This work not only deepens and extends our previous research on signal propagation but also introduces a new dimension to the understanding and application of symmetric differential equation systems, bringing the model one step closer to capturing the behavior of real biological neurons.

## 2. Symmetric differential equations and parameter choices

In the field of computational neuroscience, differential equations have long served as a foundational tool for modeling the dynamical behavior of neural systems. However, classical models such as the Hodgkin-Huxley (HH) system[12], despite their biological fidelity, are often too complex to be deployed at scale. Motivated by this challenge, we propose a novel class of symmetric differential equations grounded in symmetry principles[21, 22].

Figure 1 illustrates the conceptual development of our symmetric differential equation system. As shown in Figure 1a, we begin with the traditional *Wuxing* (Five Elements) framework from Chinese philosophy, which encodes a closed and symmetric logic of generation and inhibition among five elemental components: Metal (J), Water (S), Wood (M), Fire (H), and Earth (T). In this cyclical structure, each element either promotes (generates) or suppresses another, forming a self-contained symmetric system.

To translate this logic into a dynamic model, we reformulate the framework by incorporating self-decay and input-output nodes, while also drawing inspiration from predator-prey dynamics. The resulting system is a distributed set of differential equations, as depicted in Figure 1b and formalized in Equation (2.1). For instance, in the Wuxing logic, Earth gives rise to Metal (e.g., ore is extracted from soil), whereas Fire can melt and thereby subdue Metal. This interaction leads to the first equation in (2.1):

$dJ/dt = k_{11}T - k_{21}J - k_{31}JH$. Here, the second term represents a self-decay component, introduced to stabilize the system in analogy with the damping mechanisms in ecological equations.

Fig. 1 From Wuxing logic to symmetric differential equations

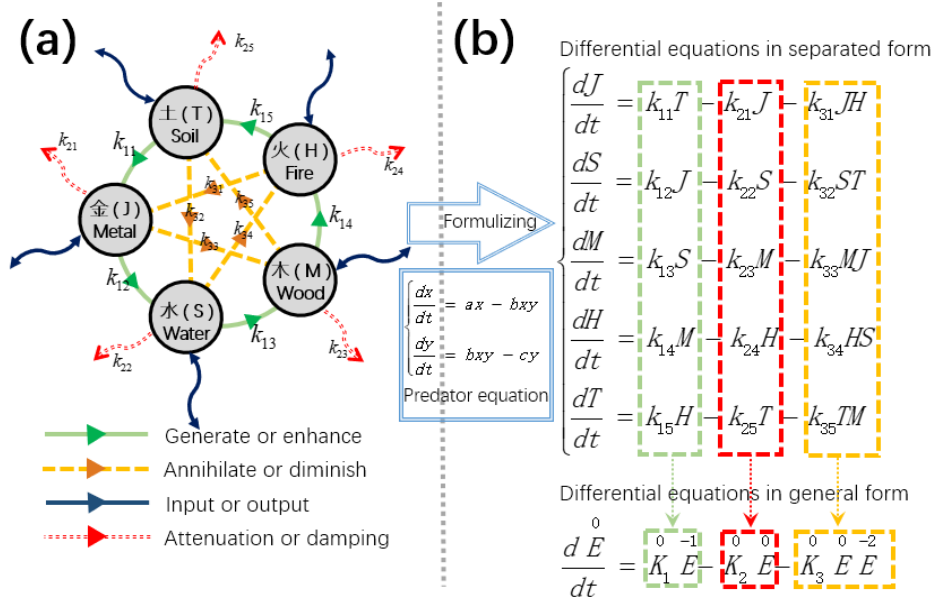

Fig. 1. a Traditional Wuxing logic posits that the world is composed of five distinct elements that interact through generative and inhibitory relationships, forming the logical framework of the universe. The logical structure depicted in the figure deviates from the traditional model by incorporating self-attenuation terms and creating interfaces for external input and output signals. In this system, there are five nodes, each capable of serving as an input or output. However, to prevent signal interference, a node can either receive input or generate output at any given time, but not both simultaneously.

Fig. 1. b By combining the Wuxing logic with the predator-prey equation, we can derive a set of differential equations. The symmetry of the system is carefully preserved throughout this transformation process. As a result, both the traditional Five Elements logic and the predator-prey equation are modified, ultimately leading to a set of fully symmetrical equations. For clarity, these equations are presented in a generalized format, with the numbers above the elements and parameters indicating the offset of each loop.

$$\begin{cases} \dfrac{dJ}{dt} = k_{11}T - k_{21}J - k_{31}JH \\ \dfrac{dS}{dt} = k_{12}J - k_{22}S - k_{32}ST \\ \dfrac{dM}{dt} = k_{13}S - k_{23}M - k_{33}MJ \\ \dfrac{dH}{dt} = k_{14}M - k_{24}H - k_{34}HS \\ \dfrac{dT}{dt} = k_{15}H - k_{25}T - k_{35}TM \end{cases} \quad (2.1)$$

Let **E** = {J, S, M, H, T} denote the set of elemental states, and define three parameter sets: $K_1 = \{k_{11}, k_{12}, k_{13}, k_{14}, k_{15}\}$, $K_2 = \{k_{21}, k_{22}, k_{23}, k_{24}, k_{25}\}$, $K_3 = \{k_{31}, k_{32}, k_{33}, k_{34}, k_{35}\}$, Then, the system in Equation (2.1) can be rewritten in a compact form as Equation (2.2).

$$\dfrac{dE}{dt} = K_1 E - K_2 E - K_3 EE \quad (2.2)$$

It is important to note that the ordering of elements within the vector **E** differs depending on the context and position in the equation, and thus cannot be trivially unified. However, if we consider a cyclic permutation of these elements across different positions, Equation (2.2) can be generalized into a rotationally symmetric form, denoted as Equation (2.3) (see Figure 1b).

$$\frac{d\overset{0}{E}}{dt} = K_1 \overset{0\ -1}{E} - K_2 \overset{0\ 0}{E} - K_3 \overset{0\ 0\ -2}{E\ E} \tag{2.3}$$

This general formulation permits arbitrary extension in the number of elements and customization of their interaction rules, provided that the cyclic structure is preserved. Consequently, the generalized form (Equation 2.3) possesses broader applicability. Due to the structural similarities among such symmetric differential systems, analytical results obtained from Equation (2.1) can naturally be extended to the broader family defined by Equation (2.3). Hence, unless otherwise specified, we will focus on the symmetric system as defined in Equation (2.1) throughout this paper.

To reduce the complexity of analysis in Equation (2.3), we assume that all parameters in $K_1$, $K_2$, and $K_3$ are strictly positive, and the initial values of all components in E are also positive. This ensures that the system state remains within the positive orthant, a property we will rigorously prove in subsequent sections. Furthermore, to guarantee the existence of a positive fixed point, we impose the condition $K_1 > K_2$.

3. System analysis

    3.1 Static Analysis

In dynamical systems, fixed point theory serves as one of the most fundamental approaches for analyzing system properties. Fixed points provide an intuitive geometric interpretation of equilibrium or steady-state solutions of differential equations. For a given differential equation, a fixed point corresponds to a state of the system where the time derivative vanishes, indicating no further change in the system's state over time. This property makes fixed points central to understanding the long-term behavior of the system. Moreover, fixed point theory plays a critical role in the qualitative analysis of differential equations; through phase space analysis, fixed points help delineate distinct behavioral regimes of the system.

In our previous studies, we defined the state at the fixed point as the *zero state* of the system, and on this basis introduced perturbation theory to characterize input and output signals.

Based on the parameter settings described in the previous section, we now investigate the fixed points and dynamical performance of the system. It is evident that $B = 0$ is a fixed point of the system; however, this point is not stable—any small perturbation will cause all variables to diverge from it. We now turn our attention to non-zero fixed points of the system.

When the system contains only one variable, the fixed point can be directly obtained through simple calculation, as shown in Equation (3.1).

$$B_1 = \frac{K_1 - K_2}{K_3} \tag{3.1}$$

Similarly, when the system involves two variables, fixed points can still be derived analytically, as illustrated in Equation (3.2):

$$\begin{cases} B_1 = \dfrac{k_{11}k_{12} - k_{21}k_{22}}{k_{12}k_{31} + k_{21}k_{32}} \\ B_2 = \dfrac{k_{11}k_{12} - k_{21}k_{22}}{k_{11}k_{32} + k_{22}k_{31}} \end{cases} \tag{3.2}$$

However, when the number of system variables exceeds three, solving for fixed points analytically becomes increasingly intractable. Nevertheless, the logic structure implied by Equation (3.1) allows us to understand and regulate the system's fixed points. Although Equation (3.1) is relatively simple, it reveals how the fixed point is modulated by the parameters $K_1$, $K_2$ and $K_3$: the fixed point increases with $K_1$, but decreases with $K_2$, and $K_3$:. In our earlier work, we proposed corresponding adjustment strategies for each of these parameters. Since such adjustments are arbitrary, the system can be configured to memorize specific information via its fixed points.

Furthermore, based on the memorized information encoded in these fixed points, the system can compare them with incoming signals and generate new responses accordingly—thus enabling a form of convolutional operation analogous to that found in convolutional neural networks (CNNs).

### 3.2 Simplified dynamic analysis

In our previous work, we implemented a training scheme for neural networks based on symmetric differential equations. In this framework, each set of symmetric differential equations is treated as an individual neural cell capable of transmitting signals and autonomously adjusting its parameters based on bidirectional signal propagation. This adjustment mechanism operates independently of traditional backpropagation algorithms, thereby laying a solid foundation for constructing biologically plausible neural systems. In this section, we conduct a stability analysis of the dynamic behavior of the symmetric differential equation system.

We begin by considering the simplest case in which the system contains only a single dynamic variable. This serves as a basis for generalizing to more complex cases. When the number of variables is one, the system reduces to a standard logistic-type equation. If the parameters satisfy the conditions outlined in Chapter 2, the system admits a positive fixed point. Specifically, by choosing $K_1 = k_{11} = 1$, $K_2 = k_{21} = 0.5$, $K_3 = k_{31} = 0.5$, we obtain Equation 3.3. The corresponding phase trajectory is illustrated in Figure 2.

$$\frac{dE_1}{dt} = 0.5 E_1 (1 - E_1) \tag{3.3}$$

Figure 2 presents a simplified view of the dynamic properties of Equation 3.3. The equation possesses two fixed points: B=0 and B=1 The origin B=0 is an unstable fixed point, whereas B=1 serves as a global attractor. When 0<E<1, the right-hand side of Equation 3.3 is positive, causing E to increase toward 1 over time. When E>1, the right-hand side becomes negative, and E decreases toward 1. For E<0, the right-hand side remains negative, resulting in divergence toward negative infinity. Since Equation 3.3 is a simplified version of Equation 2.3, similar dynamic features are also expected to emerge in the more general system described by Equation 2.3.

**Fig. 2 Time Response of the System (dE/dt=0.5*E(1-E))**

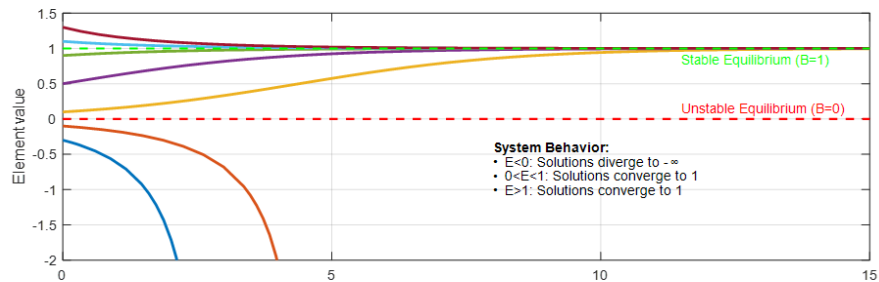

**Fig. 2.** This is the trajectory of the one-dimensional logistics equation. It can be seen that when the initial value of the element value E>0, the trajectory is attracted to the fixed point B=1, and when the initial value of the element is less than 0, the system diverges to negative infinity.

### 3.3 Comprehensive dynamic analysis

To thoroughly investigate the stability of Equation (2.3), we first index its components to explicitly reflect the different elements of the system, leading to the reformulated system described in Equation (3.4).

$$\frac{d \overset{0}{E_i}}{dt} = \overset{0}{K_1} \overset{-1}{E_{i-1}} - \overset{0}{K_2} \overset{0}{E_i} - \overset{0}{K_3} \overset{0}{E_i} \overset{-2}{E_{i-2}} \tag{3.4}$$

The stability analysis of Equation (3.4) is relatively intricate. To address it systematically, we decompose the problem into several analytical steps:

**Step 1**: Proof of Positivity

Given the initial conditions, all components of the vector E are strictly positive at t=0. Suppose that a specific component $E_i$ tends toward zero at some time. In Equation (3.4), the right-hand side corresponding to $E_i$ becomes strictly positive(Equation 3.5), primarily due to the presence of a strictly positive first-order term. This causes $E_i$ to increase and move away from zero. Therefore, once the parameters and initial values are fixed, all components of Equation (3.4) remain strictly positive for t>0. Furthermore, each $E_i$ admits a positive lower bound M1>0, which may be arbitrarily small but not infinitesimal.

$$\frac{d \overset{0}{E_i}}{dt} = \overset{0}{K_1} \overset{-1}{E_{i-1}} - \overset{0}{K_2} \overset{0}{E_i} - \overset{0}{K_3} \overset{0}{E_i} \overset{-2}{E_{i-2}} \approx \overset{0}{K_1} \overset{-1}{E_{i-1}} > 0 \tag{3.5}$$

**Step 2**: Proof of Boundedness

To establish boundedness, we construct a Lyapunov function based on Lyapunov's second method:

$$V(t) = \sum_i^{En} E_i = J + S + M + H + T \tag{3.6}$$

where $E_n$ represent the system's element number, and V(t)>0 for all t>0 by Step 1.Taking the time derivative of V(t), we can get:

$$\frac{dV(t)}{dt} = \sum_i^{En} \overset{0}{K_1} \overset{-1}{E_{i-1}} - \sum_i^{En} \overset{0}{K_2} \overset{0}{E_i} - \sum_i^{En} \overset{0}{K_3} \overset{0}{E_i} \overset{-2}{E_{i-2}} \tag{3.7}$$

and regrouping terms, we obtain:

$$\frac{dV(t)}{dt} = L(E_i) - Q(E_i) \tag{3.8}$$

where:

$$L(E_i) = (k_{12} - k_{21})J + (k_{13} - k_{22})S + (k_{14} - k_{23})M + (k_{15} - k_{24})H + (k_{11} - k_{25})T \tag{3.9}$$

and:

$$Q(E_i) = k_{31}JH + k_{32}ST + k_{33}MJ + k_{34}HS + k_{35}TM \tag{3.10}$$

Here, $L(E_i)$ denotes the linear terms and $Q(E_i)$ the quadratic terms associated with each $E_i$. Based on the positivity established in Step 1, it is always possible to find sufficiently large values of $E_i$ such that the quadratic terms dominate and ensure $dV(t)/dt < 0$. This implies that the system described by Equation (3.4) is bounded within the positive domain.

**Step 3:** Proof of the Existence of a Fixed Point within a Bounded Region

Consider an n-dimensional continuous dynamical system $dx/dt = f(x)$，If, for t>0, all trajectories of the system are confined within a bounded region Ω, where Ω is a compact and convex set, then it follows that the equation $f(x) = 0$ has at least one solution in Ω; that is, the system admits at least one fixed point within $\Omega$.

We now define a mapping T: $\Omega \to \Omega$ as follows:
$$T(x) = x + \varepsilon f(x) \tag{3.11}$$

Where $\varepsilon > 0$ is a small parameter to be determined.

Our goal is to verify the existence of a sufficiently small $\varepsilon > 0$ such that $T(x) \in \Omega$ for all $x \in \Omega$.

Note that f(x) is a continuous function and $\Omega$ is a bounded closed set, which implies that f(x) is bounded on $\Omega$. Hence, for any $x \in \Omega$, there exists a constant M>0 such that:
$$M := \sup_{x \in \Omega} \|f(x)\| < \infty \tag{3.12}$$

for any $x \in \Omega$, such that
$$\|T(x) - x\| = \|\varepsilon f(x)\| \leq \varepsilon M \tag{3.13}$$

We can choose $\varepsilon > 0$ sufficiently small such that:
$$\varepsilon M < \frac{d(\partial \Omega, \text{in}(\Omega))}{2} = \frac{\delta}{2} \tag{3.14}$$

where $\delta$ is the distance from the boundary of $\Omega$ to ensure that T(x) remains strictly inside $\Omega$. This guarantees that t $T(x) \in \Omega$ for all $x \in \Omega$.

Thus, we have a continuous mapping $T: \Omega \to \Omega$, where $\Omega$ is a non-empty, compact, and convex subset of $R^n$. By Brouwer's Fixed Point Theorem, there exists a point $x^* \in \Omega$ such that:
$$T(x^*) = x^* \tag{3.15}$$

From the definition of $T$, it follows that:
$$x^* + \varepsilon f(x^*) = x^* \Rightarrow \varepsilon f(x^*) = 0 \Rightarrow f(x^*) = 0 \tag{3.16}$$

This confirms that the equation f(x)=0 admits at least one solution within $\Omega$ and the proposition is proved.

**Step 4:** Proof of the Uniqueness of the Fixed Point of Equation (3.4)

We begin by approximating the fixed point of Equation (3.4) using the form given in Equation (3.17).
$$B \approx \frac{\overline{K_1} - \overline{K_2}}{\overline{K_3}} \tag{3.17}$$

In this expression, $\overline{K_1}$ $\overline{K_2}$ $\overline{K_3}$, denote the average values of the three groups of parameters described earlier. When the parameters within each group are identical, Equation (3.17) provides an exact solution. Although Equation (3.17) only gives an approximate fixed point of Equation (3.4), it reveals the qualitative relationship between the fixed point B and the parameters $K_1$, $K_2$ and $K_3$:. Specifically, B is positively correlated with $K_1$; as $K_1$ increases monotonically, so does B. Conversely, B is negatively correlated with $K_2$ and $K_3$; an increase in either $K_2$ or $K_3$ results in a monotonic decrease in B. It is important to note that this monotonic relationship is local. The fixed point B is a 5-dimensional vector, and each component $B_i$ is strictly positively correlated with $k_{1i}$ and strictly negatively correlated with $k_{2i}$ and $k_{3i}$.

let:
$$k_1^{max} = \max\{k_{11},\ k_{12},\ k_{13},\ k_{14},\ k_{15}\} \tag{3.18}$$

$$k_1^{min} = \min\{k_{11}, \ k_{12}, \ k_{13}, \ k_{14}, \ k_{15}\} \tag{3.19}$$

$$k_2^{max} = \max\{k_{21}, \ k_{22}, \ k_{23}, \ k_{24}, \ k_{25}\} \tag{3.20}$$

$$k_2^{min} = \min\{k_{21}, \ k_{22}, \ k_{23}, \ k_{24}, \ k_{25}\} \tag{3.21}$$

$$k_3^{max} = \max\{k_{31}, \ k_{32}, \ k_{33}, \ k_{34}, \ k_{35}\} \tag{3.22}$$

$$k_3^{min} = \min\{k_{31}, \ k_{32}, \ k_{33}, \ k_{34}, \ k_{35}\} \tag{3.23}$$

We now introduce two auxiliary equations, (3.24) and (3.25)

$$\frac{d \overset{0}{E_i}}{dt} = k_1^{max} \overset{-1}{E_{i-1}} - k_2^{min} \overset{0}{E_i} - k_3^{min} \overset{0}{E_i} \overset{-2}{E_{i-2}} \tag{3.24}$$

$$\frac{d \overset{0}{E_i}}{dt} = k_1^{min} \overset{-1}{E_{i-1}} - k_2^{max} \overset{0}{E_i} - k_3^{max} \overset{0}{E_i} \overset{-2}{E_{i-2}} \tag{3.25}$$

In (3.24) and (3.25) all parameters within each group are set to be identical. These equations respectively yield two fixed points, $B_{max}$ and $B_{min}$. Due to the symmetry of the system, $B_{max}$ and $B_{min}$ are the unique fixed points of their respective equations. Since Equation (3.17) provides a good approximation of the fixed point of Equation (3.4), it follows that the actual fixed point of Equation (3.4) lies between $B_{max}$ and $B_{min}$.

$$B_{min} < B < B_{max} \tag{3.26}$$

To formalize the transition between these two extremes, we define a set of interpolating equations (Equation 3.27),

$$\frac{d \overset{0}{E_i}}{dt} = k_1^{S_1} \overset{-1}{E_{i-1}} - k_2^{S_2} \overset{0}{E_i} - k_3^{S_3} \overset{0}{E_i} \overset{-2}{E_{i-2}} \tag{3.27}$$

where:

$$k_1^{S_1} = (1-S_1)k_1^{max} + S_1 k_2^{min} \tag{3.28}$$

$$k_2^{S_2} = (1-S_2)k_2^{min} + S_2 k_2^{max} \tag{3.29}$$

$$k_3^{S_3} = (1-S_3)k_3^{min} + S_3 k_3^{max} \tag{3.30}$$

where $S_1$, $S_2$, $S_3 \in [0,1]^5$. By continuously varying $S_1$, $S_2$, $S_3$, one can smoothly transition from Equation (3.24) to Equation (3.25). This construction allows us to traverse the entire state space between the two extreme cases, and the state space corresponding to Equation (3.4) must be included in this transition.

According to the previous monotonicity results, when the parameters $K_1$, $K_2$ and $K_3$ change monotonically, the corresponding fixed point B also changes monotonically. Although this monotonicity is conditional, a properly chosen sequence of parameter adjustments ensures that the fixed point remains unique throughout the transformation path. For instance, one can first adjust the first parameter in $K_1$, maintaining monotonicity, then adjust the second, and so on. By carefully regulating the full set of parameters in $K_1$, $K_2$ and $K_3$, we guarantee that the fixed point evolves in a strictly monotonic manner along the transition path. As a result, no bifurcations occur, and the uniqueness of the fixed point in Equation (3.4) is preserved throughout.

**Step 5:** Analysis of the Dynamical Behavior of the System

The dynamical behavior of the system is analyzed in two parts. First, if the trajectory of the system terminates at a fixed point, then the ω-limit set contains this fixed point, which is a natural property of the system. Second, if the trajectory does not terminate at the fixed point, it will return to a neighborhood of the fixed point infinitely often.

To study the local dynamics of Equation (3.4) near the fixed point, we decompose the system state $E(t)$ into two components: B and D, where B denotes the fixed point of the system and D is the perturbation component, i.e., $E(t) = B + D$. According to previous derivations, both E and B are strictly positive, whereas the sign of D is generally unknown. We now discuss two possible scenarios:

**Case 1:** Some elements of D are positive while others are negative, and this sign pattern persists over time.

In Equation (3.4), there exist two cyclic structures: the first term represents a generative loop among the variables （$J \to S \to M \to H \to T \to J$）, while the third term encodes a suppressive loop ($J \to M \to T \to S \to H \to J$). If the components of D remain of mixed signs, the system's cyclic interactions lead to alternating effects, causing each component to potentially cross zero at some point. However, since this zero-crossing occurs for individual variables at different times, it follows that the system trajectory passes through a neighborhood of the fixed point rather than the fixed point itself.

**Case 2:** After some time t2, all components of D become either strictly positive or strictly negative and remain so thereafter. In this case, the system trajectory would cease to revisit the neighborhood of the fixed point. We now show that such a situation is impossible.

Using a proof by contradiction, suppose that for t>t2>0, all elements of D satisfy D>0. Given that D>0 and the system remains bounded in the positive orthant, we can construct a new convex region that excludes the original fixed point B. By repeating the arguments used in Step 3, it can be shown that a new fixed point $B_{new}$ must exist within this new convex region. However, this contradicts the previous conclusion that the system has a unique fixed point in the positive space. Therefore, this scenario cannot occur, and it must be that the system trajectory returns to the neighborhood of the fixed point infinitely often.

In conclusion, we have completed the stability analysis of Equation (3.4). The main findings are as follows: the system is bounded in the positive domain, and there exists a unique fixed point within this bounded region. If the system trajectory terminates at the fixed point, then the ω-limit set reduces to a singleton, indicating that the system is asymptotically stable. On the other hand, if the trajectory does not terminate at the fixed point but revisits its neighborhood infinitely often, the system is Lyapunov stable.

4. System Functionality and Simulation

In Chapter 3, we conducted a rigorous mathematical analysis demonstrating that the proposed symmetric differential equation system possesses a functional duality: depending on specific parameter settings and system states, its trajectories may either asymptotically stabilize at a fixed point or exhibit sustained oscillations near that point. In our previous work, we utilized the system in its stable mode as a signal propagator, enabling the successful training of a large-scale neural network via reverse signal propagation. This approach achieved reliable performance on the MNIST dataset. In the present study, we shift focus to exploring the system's role as a signal generator.

4.1 First Type of Signal Generator

Our analysis indicates that in order for the system to function as a signal generator, there must exist an unbalanced feedback loop within the system. This imbalance leads to persistent oscillatory behavior and the emergence of periodic signal patterns. In Equation 3.4, two feedback loops are present: the first term reflects the generative logic of system components, while the third term captures the inhibitory interactions. These two loops interact to establish the basis for signal propagation. To ensure system stability, a second term is introduced to Equation 3.4, acting as a damping component. Consequently, by adjusting the coefficient K2 associated with the second term, one can effectively modulate the convergence domain of the system and thus control the emergence of oscillatory dynamics.

Based on Equation 3.4, we constructed two distinct models using different parameter settings and performed numerical simulations (Figure 3) to illustrate the system's signal-generating capabilities.

**Fig. 3 Time Response of Two Systems with Different Initial Value**

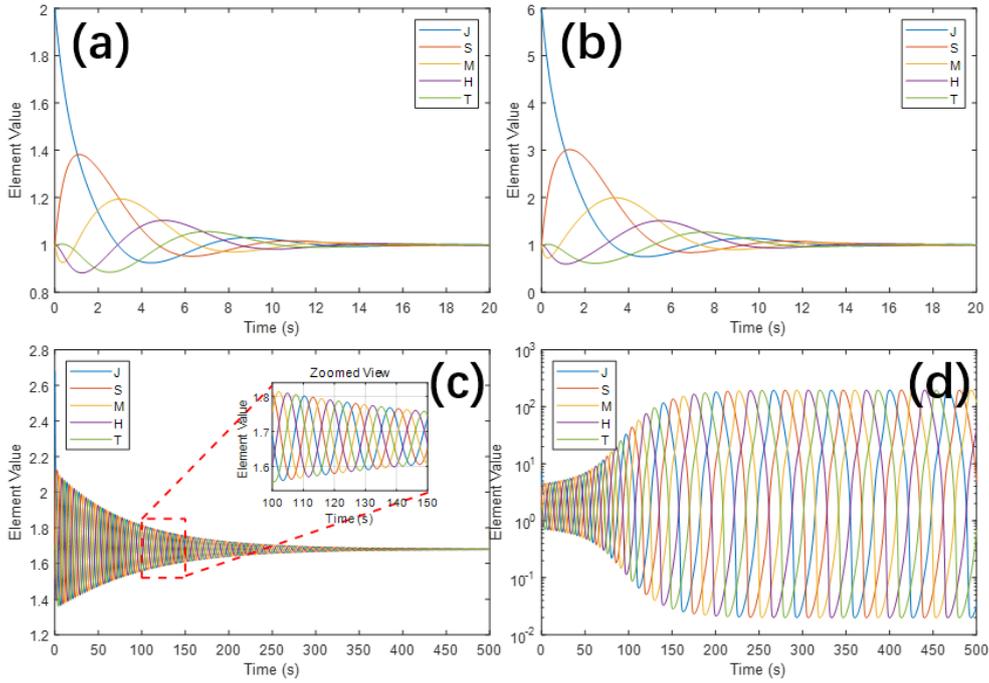

**Fig. 3. a.** Simulation results based on Equation (3.4), using a relatively large value of $K_2$ and a small initial deviation from the fixed point. The system quickly returns to the fixed point, demonstrating strong stability under these conditions.

**Fig. 3. b.** This simulation uses the same $K_2$ value as in Figure 3a but with a larger initial deviation. Compared to Figure 3a, although the response amplitude increases due to the larger deviation, the system exhibits similar dynamic characteristics and still converges to the fixed point.

**Fig. 3. c.** Simulation results based on Equation (3.4) with a smaller $K_2$ and a small initial deviation. The reduced $K_2$ weakens the system's stability, leading to prolonged oscillations even with minor perturbations. However, the system eventually returns to the fixed point.

**Fig. 3. d.** This simulation uses the same small $K_2$ value as in Figure 3c but with a larger initial deviation. The system enters sustained oscillations and fails to return to the original fixed point. A logarithmic scale is used in this figure to better visualize the dynamic behavior due to the larger amplitude of oscillations.

Figure 3 illustrates the dynamic responses of the system under different configurations of the $K_2$ parameter. In Figures 3a and 3b, the system parameters are set as $K_1$={1, 1, 1, 1, 1}; $K_2$={0.5, 0.5, 0.5,

0.5, 0.5}; and $K_3$={0.5, 0.5, 0.5, 0.5, 0.5}. At time t=0, the initial conditions are E={2,1,1,1,1} in Figure 3a, and E={6,1,1,1,1} in Figure 3b. As observed, the dynamic behaviors in both figures are qualitatively similar, differing primarily in amplitude. This is mainly due to the different initial deviations from the fixed point. However, in both cases, the system quickly returns to the equilibrium state as time progresses.

In Figures 3c and 3d, a smaller $K_2$ value is adopted, specifically $K_2$={0.16,0.16,0.16,0.16,0.16}, while keeping $K_1$ and $K_3$ unchanged. The initial conditions at t=0 are set as E = {2.68,1.68,1.68,1.68,1.68} for Figure 3c and E = {6.68,1.68,1.68,1.68,1.68} for Figure 3d. In Figure 3c, the system exhibits frequent oscillations. Although all components eventually return to the fixed point, the convergence takes significantly longer—approximately 20 times slower than in Figure 3a. It is worth noting that the fixed point has shifted to B=1.68 due to the new $K_2$ values. To maintain a consistent initial deviation relative to the fixed point, the initial values have been adjusted accordingly.

In Figure 3d, due to the reduced convergence basin of the system, a larger initial deviation leads to persistent periodic oscillations. The system fails to return to its equilibrium and instead forms a stable, closed loop in the phase space. It should be emphasized that a logarithmic scale is used in Figure 3d due to the wide range of oscillation amplitudes. As shown, while the minimum values of the signals can become very small, they do not decay to zero but stabilize around a finite value—consistent with our earlier theoretical analysis.

These results demonstrate that by tuning the $K_2$ parameter, the system can flexibly switch between different functional modes. With a larger $K_2$, the system functions as a signal propagator: upon receiving an input signal, it transmits the signal and rapidly returns to the fixed point once the input ceases. Conversely, reducing $K_2$ renders the system prone to oscillations, making it function as a signal generator that continuously produces oscillatory outputs. Through appropriate parameter design, desired oscillatory dynamics can be achieved.

4.2 Second Type of Signal Generator

In the previous section, we demonstrated that by adjusting the parameter K2, one can smoothly control the convergence domain of the system, thereby enabling the system to function either as a signal propagator or as a signal generator. However, tuning system behavior is not limited to parameter modulation. In this section, we introduce a second approach: altering the system's structure to induce a functional change.

Specifically, we constructed two distinct system configurations, as illustrated in Figure 4.

The system structure in Figure 4a is described by Equation (4.1),

$$\frac{d E_{En=8}^{0}}{dt} = K_1^0 E_{En=8}^{-1} - K_2^0 E_{En=8}^{0} - K_3^0 E_{En=8}^{0} E_{En=8}^{-2} \tag{4.1}$$

While the system structure in Figure 4b is defined by Equation (4.2).

$$\frac{d E_{En=8}^{0}}{dt} = K_1^0 E_{En=8}^{-1} - K_2^0 E_{En=8}^{0} - K_3^0 E_{En=8}^{0} E_{En=8}^{-3} \tag{4.2}$$

In symmetric differential systems, two types of interaction loops coexist: the generative loop and the inhibitory loop. If we take the generative loop as a structural reference, it is always possible to construct a generative cycle with an offset of -1. By modifying the offset within the inhibitory loop, different system equations can be obtained. In both figures, the system contains 8 elements; the key difference lies in the inhibitory loop offset: -2 in Figure 4a and -3 in Figure 4b.

**Fig. 4 Two Systems with 8 Elements of Different Structures**

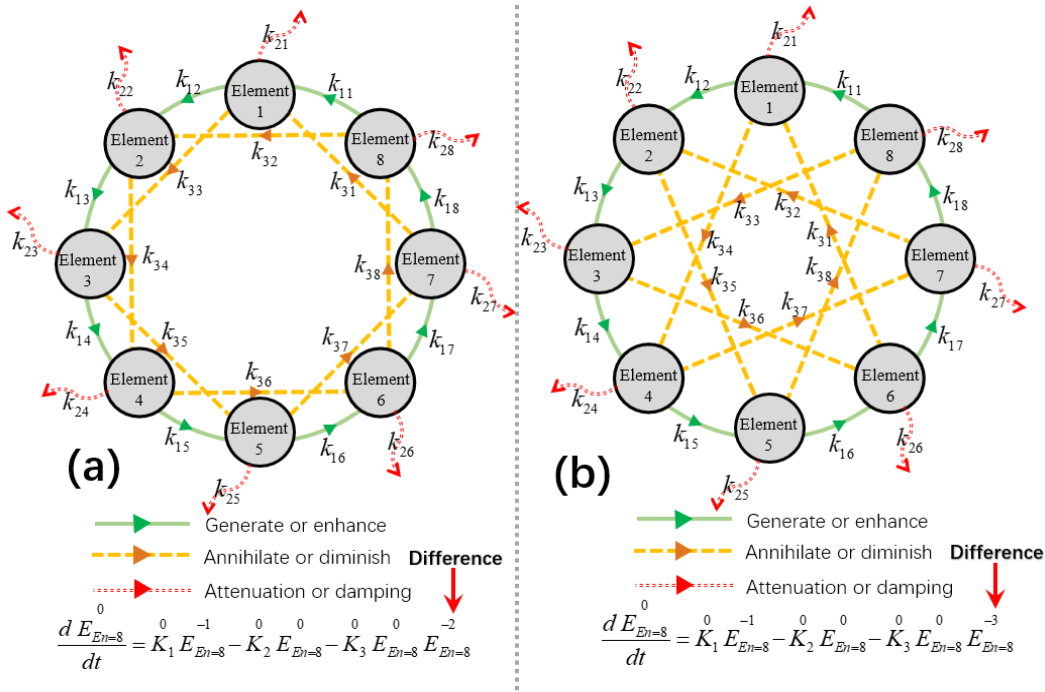

**Fig. 4. a.** Symmetric differential equation structure with 8 elements. The deviation parameter for the generative loop is set to –1, and for the suppressive loop it is –2. When the signal enters from Element 1, it passes through Element 2 and Element 3. The generative and suppressive loops first intersect at this point, reaching a temporary dynamic equilibrium.

**Fig. 4. b.** Symmetric differential equation structure with 8 elements. The deviation parameter for the generative loop remains –1, while that for the suppressive loop is increased to –3. Compared to Fig. 4a, the generative and suppressive loops are now further apart. When the signal enters from Element 1, it must traverse Element 2, Element 3, and Element 4 before reaching the first temporary equilibrium.

When an input signal is applied to Element 1, the system state deviates from its fixed point. According to our design, this deviation is then treated as the output signal corresponding to the input. However, for the system to remain stable, it is crucial that the state eventually returns to the fixed point. In the system of Figure 4a, the imbalance must propagate through Elements 2 and 3 before encountering the inhibitory feedback initiated by Element 1, thereby achieving the first transient balance. In contrast, in the system of Figure 4b, the imbalance must traverse a longer path—Elements 2, 3, and 4—before receiving any inhibitory feedback from Element 1. This extended delay to reach the initial transient balance implies that the system in Figure 4b is more susceptible to oscillation than that in Figure 4a.

Figure 5 presents simulation results for both systems under various initial conditions. Figures 5a and 5b are based on Equation (4.1), whereas Figures 5c and 5d correspond to Equation (4.2). All simulations in Figure 5 use identical parameters:

$K_1=\{1,1,1,1,1,1,1,1\}, K_2=\{0.5,0.5,0.5,0.5,0.5,0.5,0.5,0.5\}, K_3=\{0.5,0.5,0.5,0.5,0.5,0.5,0.5,0.5\}$.

The initial conditions at t=0 are as follows:

Figure 5a: E=\{2,1,1,1,1,1,1,1\}

Figure 5b: E=\{6,1,1,1,1,1,1,1\}

Figure 5c: E=\{1.01,1,1,1,1,1,1,1\}

Figure 5d: E={6,1,1,1,1,1,1,1}

**Fig. 5 Time Response of Two Systems with 8 Elements of Different Structures**

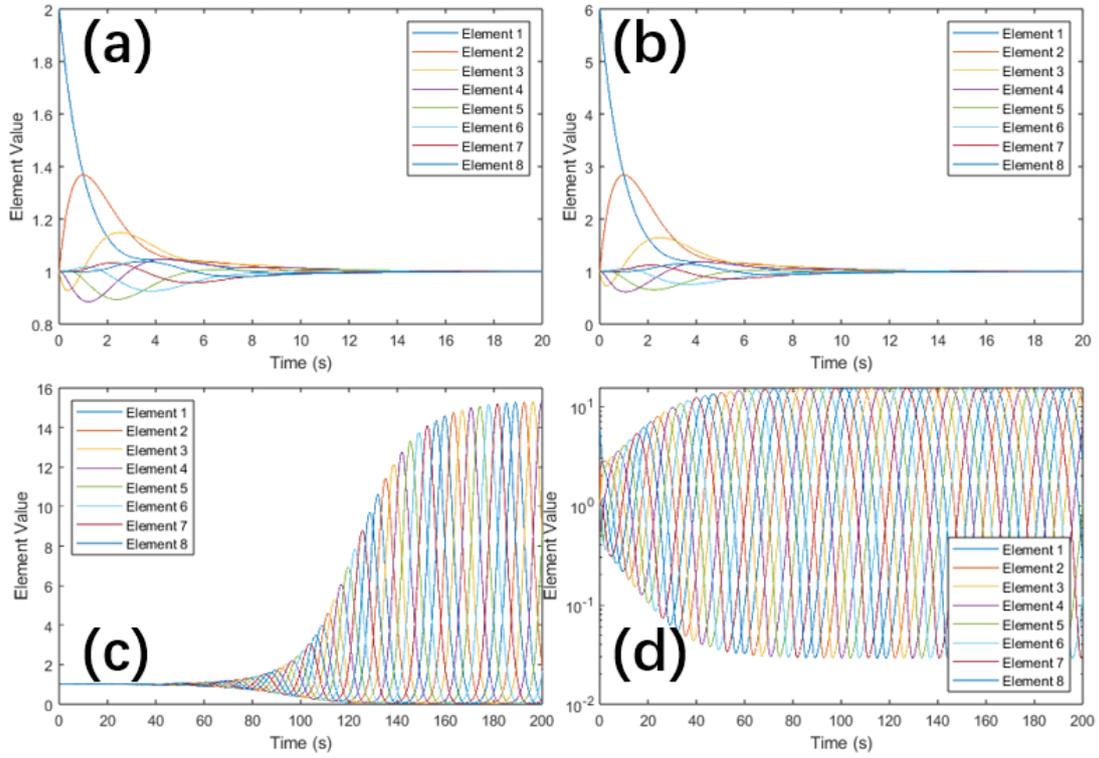

**Fig. 5. a.** Simulation result based on Equation (4.1). When the initial state slightly deviates from the fixed point, the system quickly returns to the original fixed point. The result is very similar to that in Figure 3a.

**Fig. 5. b.** Simulation result based on Equation (4.1). Even with a larger initial deviation from the fixed point, the system still returns to the original fixed point in a short time. Apart from a difference in amplitude, the dynamic behavior is nearly identical to that shown in Figure 5a.

**Fig. 5. c.** Simulation result based on Equation (4.2). When the initial deviation is extremely small, the system starts to exhibit noticeable oscillatory behavior after approximately 60 seconds.

**Fig. 5. d.** Simulation result based on Equation (4.2). With a larger initial deviation, the system enters an oscillatory state more rapidly. However, the final oscillation amplitude is identical to that in Figure 5c, indicating that the oscillation is an inherent property of the system.

As shown in Figures 5a and 5b, the system quickly returns to its fixed point, exhibiting behavior similar to that observed in Figure 3. In contrast, even a small initial deviation in Figure 5c leads to significant oscillations after approximately 60 seconds. Figure 5d shows that with a larger initial deviation, the system enters an oscillatory regime immediately, reaching the same oscillation amplitude as in Figure 5c. This indicates that oscillatory behavior is an intrinsic property of the system in Figure 4b.

Figure 5 demonstrates that even under identical parameter settings, modifying the internal connectivity structure can fundamentally alter the system's operational state. Compared to parameter adjustments such as changing $K_2$, structural changes produce a more dramatic and immediate shift in system dynamics. Therefore, the choice of modulation method—either structural or parametric—should be determined by the specific application requirements.

## 4.3 On-road Energy

In our previous studies, we investigated the oscillatory states of the system, which are all periodic in nature and depend solely on the system's internal parameters and structure. As a result, the asymptotic behavior of each individual component tends to exhibit similar patterns. Although phase portraits and time series can intuitively illustrate the system's final states—whether stable or oscillatory—they fall short in scenarios where a real-time, quantitative measure of the system's long-term evolution is needed. To address this gap, we propose a novel intermediate-state observable termed "On-road Energy". This quantity is designed to capture the dynamic energy inherent in the system enroute to its final state. The central idea is: if the on-road energy monotonically decays to zero over time, it indicates efficient energy dissipation within the system, ultimately stabilizing at the lowest-energy fixed point. Conversely, if the on-road energy maintains a nonzero level or exhibits persistent fluctuations, it implies a balance between energy dissipation and replenishment, and the system will exhibit sustained oscillations.

To formalize this, we decompose the system's state variable E(t) as follows:

$$E(t) = B + D(t) \tag{4.3}$$

where B denotes the system's fixed point, and D(t) represents the deviation from equilibrium—capturing the dynamic, non-equilibrium component of the system. Substituting Equation (4.3) into Equation (3.4), we derive Equation (4.4):

$$\begin{aligned}
\frac{d \overset{0}{E_i}}{dt} &= \overset{0}{K_1} \overset{-1}{E_{i-1}}(t) - \overset{0}{K_2} \overset{0}{E_i}(t) - \overset{0}{K_3} \overset{0}{E_i}(t) \overset{-2}{E_{i-2}}(t) \\
&= \overset{0}{K_1} \overset{-1}{B_{i-1}} - \overset{0}{K_2} \overset{0}{B_i} - \overset{0}{K_3} \overset{0}{B_i} \overset{-2}{B_{i-2}} \\
&\quad + \overset{0}{K_1} \overset{-1}{D_{i-1}}(t) - \overset{0}{K_2} \overset{0}{D_i}(t) - \overset{0}{K_3} \overset{0}{D_i}(t) \overset{-2}{B_{i-2}} - \overset{0}{K_3} \overset{0}{B_i} \overset{-2}{D_{i-2}}(t) \\
&\quad - \overset{0}{K_3} \overset{0}{D_i}(t) \overset{-2}{D_{i-2}}(t) \\
&= F(B) + L(D) + Q(D)
\end{aligned} \tag{4.4}$$

where:

$$F(B) = \overset{0}{K_1} \overset{-1}{B_{i-1}} - \overset{0}{K_2} \overset{0}{B_i} - \overset{0}{K_3} \overset{0}{B_i} \overset{-2}{B_{i-2}} \tag{4.5}$$

$$L(D) = \overset{0}{K_1} \overset{-1}{D_{i-1}}(t) - \overset{0}{K_2} \overset{0}{D_i}(t) - \overset{0}{K_3} \overset{0}{D_i}(t) \overset{-2}{B_{i-2}} - \overset{0}{K_3} \overset{0}{B_i} \overset{-2}{D_{i-2}}(t) \tag{4.6}$$

$$Q(D) = \overset{0}{K_3} \overset{0}{D_i}(t) \overset{-2}{D_{i-2}}(t) \tag{4.7}$$

Equation 4.4 separates the system dynamics into 3 parts: Equation (4.5) reflects the equilibrium condition, with F(B)=0 by definition; Equation (4.6) accounts for the linear evolution of D(t); Equation (4.7) describes the nonlinear interactions in D(t) encapsulating the energy variation induced by system imbalance.

Based on Equation (4.7), we define the on-road energy, denoted as R(D), which serves as a quantitative measure of the transient dynamics. This indicator is closely tied to the system's perturbation structure and thus offers a compact yet powerful tool for predicting the system's evolutionary tendencies.

$$R(D) = \sum_{i}^{En} \overset{0}{K_3} \overset{0}{D_i}(t) \overset{-2}{D_{i-2}}(t) \tag{4.8}$$

**Fig. 6 Element Value and On-road Energy Based on Fig.5**

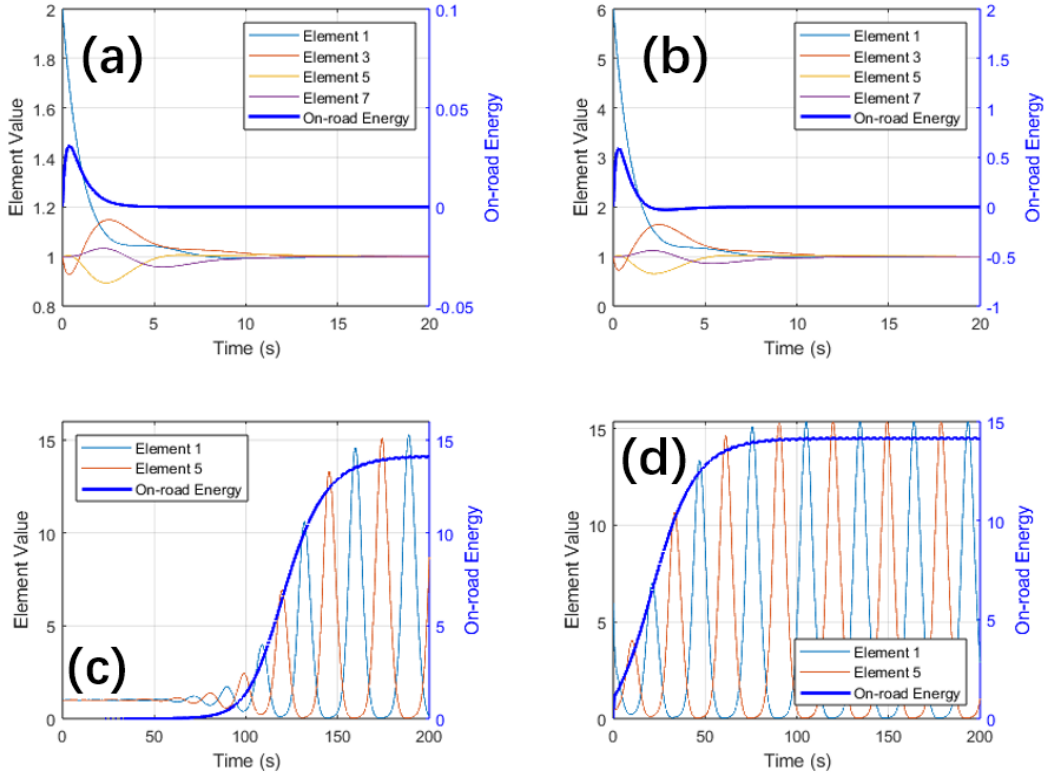

**Fig. 6. a.** Time evolution comparison between element values and On-road Energy based on the system in Fig. 5a. Compared to the complex and fluctuating trajectories of the element values, the On-road Energy curve exhibits a clear downward trend at an early stage. This indicates that On-road Energy is a fast and effective indicator for judging the convergence behavior of the system.

**Fig. 6. b.** Time evolution comparison based on Fig. 5b. The overall trend is consistent with Fig. 6a. Even when the element values are still oscillating, the On-road Energy curve clearly signals that the system is approaching a stable state.

**Fig. 6. c.** Time evolution comparison based on Fig. 5c. The continuous rise of the On-road Energy is clearly observed, matching the sustained oscillations in the system. However, as a global variable, On-road Energy more comprehensively reflects the overall level of dynamical fluctuations in the system.

**Fig. 6. d.** Time evolution comparison based on Fig. 5d. The On-road Energy curve rises rapidly, aligning with the oscillatory behavior of the system.

Figure 6 illustrates the temporal evolution of both component values (from Figure 5) and the corresponding on-road energy. For clarity, only a subset of variables from Figure 5 is shown, with a dual-Y axis configuration: the left Y-axis represents the magnitude of the system components, while the right Y-axis corresponds to the magnitude of the on-road energy. Comparing the time trajectories of on-road energy in both convergent and oscillatory regimes provides valuable insight into the system's global behavior.

As shown in Figures 6a and 6b, even in the early stages of system evolution—when individual components are still oscillating—the on-road energy demonstrates a marked downward trend. This suggests that monitoring the on-road energy allows for earlier and more reliable detection of system convergence than direct observation of component values. In contrast, Figures 6c and 6d show on-road energy increasing and eventually stabilizing, in accordance with the system's persistent oscillatory behavior. Notably, the peak value of on-road energy does not exactly match the amplitude of the system's oscillations, indicating that a portion of non-oscillatory energy continues to circulate within the system in a balanced manner.

In summary, as a global variable, on-road energy effectively captures the dynamic energy associated with oscillatory modes, helping to prevent misjudgments of system convergence that may arise from temporary fluctuations in individual components. The use of on-road energy thus enables both accurate and efficient identification of the system's long-term behavior.

4.4 Signal Suppression

A powerful neuron model should not only be capable of simulating intrinsic biological activities but also respond to and be regulated by external inputs. Having demonstrated that the system can generate self-sustained oscillations, a natural follow-up question arises: Can these oscillations be effectively suppressed or controlled? Biologically, this corresponds to the modulation of neuronal rhythmic activity by external stimuli, while in engineering terms, it relates to system stability and control. This section investigates how the introduction of an external inhibitory signal can be used to actively "quench" the oscillatory behavior of the system. Specifically, we apply a sufficiently strong external perturbation to stabilize the system at a new equilibrium point.

As illustrated in Figure 7, the system exhibits stable oscillatory dynamics in the absence of external intervention. At time $t=t_0=250s$, an inhibitory signal is introduced. We observe that the oscillation amplitude rapidly decays, and the system transitions successfully to a new stable fixed point. This result further supports the biological plausibility of the model and suggests its potential for applications in neuromorphic circuits where precise initiation and cessation of rhythmic activity are required, such as in artificial central pattern generators.

In Figure 7, we examine two previously established oscillatory models (refer to Figures 3d and 5d). The model in Figure 3d achieves oscillation by adjusting the parameter K2, while the model in Figure 5d relies on structural changes in the system's connectivity. Figures 7a and 7b are based on the model from Figure 3d. When $t>t_0=250s$, we apply an external input signal to one of the model's components. In Figure 7a, the input signal has an amplitude of 0.5; in Figure 7b, the amplitude is increased to 1.5.

As shown in Figure 7a, the smaller external input fails to stabilize the system; instead, it introduces new oscillations. Analysis of the on-road energy reveals that, in the early phase, the system exhibits short-period intrinsic oscillations. After the external signal is applied, a longer-period oscillation emerges, indicating that the short-period behavior is inherent to the system, while the longer-period oscillation is induced by the external input.

In contrast, Figure 7b shows that a stronger input successfully drives the system to a new stable state. The on-road energy rapidly declines to zero, which aligns with theoretical expectations.

Figures 7c and 7d are based on the model from Figure 5d. Similarly, when $t>t_0=250s$, we introduce an external input to one of the components—0.5 in Figure 7c and 1.0 in Figure 7d. The results are consistent with those in Figures 7a and 7b: smaller inputs fail to stabilize the system, while sufficiently large inputs lead to a new steady state. In Figure 7d, the on-road energy briefly spikes due to the system's transition to a different fixed point, followed by a rapid decay to zero, indicating convergence to the new equilibrium.

The results in Figure 7 collectively demonstrate that both types of oscillatory signal generators can be suppressed by applying external signals of adequate strength. This confirms the controllability of the system and further enhances the model's applicability in biologically inspired control scenarios.

**Fig. 7 Stabilize the System at a New Fixed Point through External Intervention**

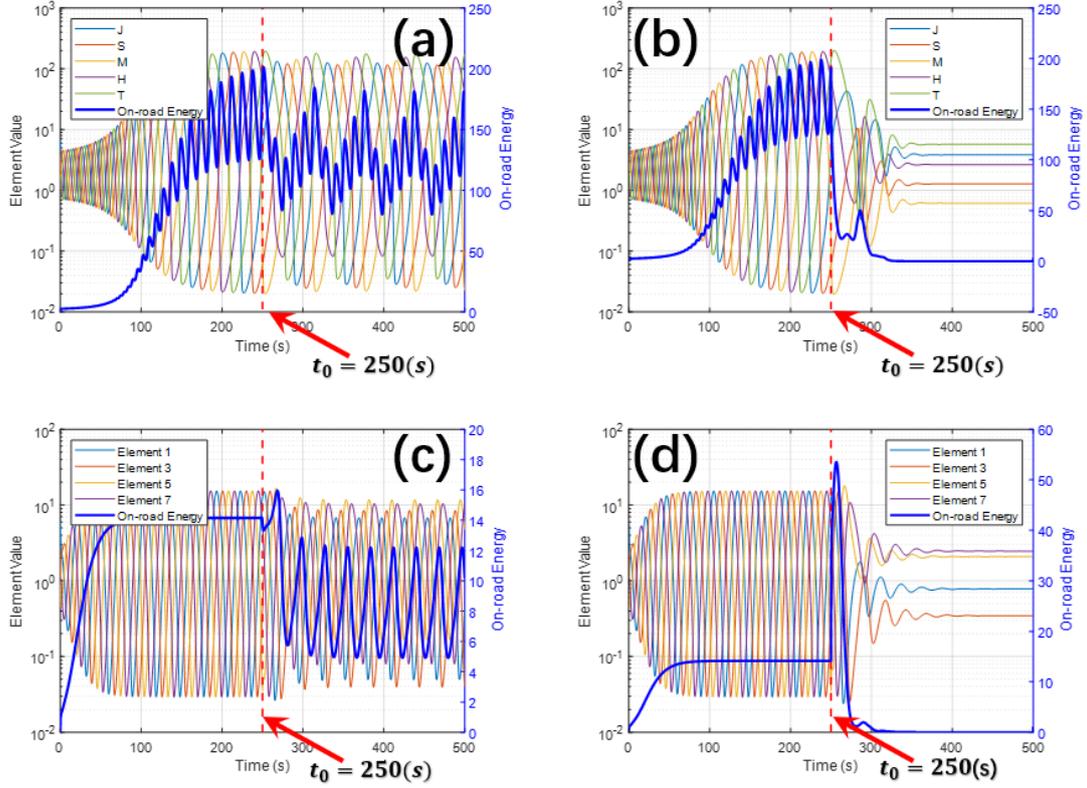

**Fig. 7. a.** In the model presented in Fig. 3d, the injection of a small external signal leads to a partial suppression of oscillations. However, the system fails to stabilize at a new fixed point. According to Equation 4.8, the computed On-road energy continues to exhibit oscillatory behavior. Two distinct oscillation periods are observed: a shorter intrinsic period that exists from the initial stage, and a longer period induced by the injected signal. This suggests that while the system possesses an inherent oscillatory mode, the external input introduces an additional slower rhythm.

**Fig. 7. b.** When a larger external signal is applied to the model in Fig. 3d, the system rapidly converges to a new fixed point. Simultaneously, the On-road energy quickly decays to zero, indicating that the system has successfully transitioned into a new stable state.

**Fig. 7. c.** In the model from Fig. 5d, a small external signal also reduces the amplitude of oscillations but is insufficient to drive the system into a stable state. The On-road energy remains oscillatory after signal injection, similar to the behavior observed in Fig. 7a. This highlights that a weak external signal is inadequate for stabilizing the system.

**Fig. 7. d.** Upon injecting a strong external signal into the model in Fig. 5d, the system promptly transitions into a stable regime. Due to the shift in the system's fixed point, the On-road energy computed via Equation 4.8 exhibits a prominent spike shortly after signal injection, followed by a rapid decay to zero. This pattern marks the system's entry into a stable state.

5. Conclusion

In this work, we have systematically investigated the dynamic characteristics of a symmetric differential equation-based neural model. We proved that the system's state space is bounded, and that its trajectories either converge to a unique positive fixed point or recurrently return to its neighborhood. This property reveals the model's potential to function not only as a signal propagator but also as a signal generator under different conditions.

Numerical simulations further confirm the existence of two distinct modes through which the system can generate periodic oscillations. The first method involves adjusting the parameter $K_2$, which allows a balanced modulation of the system's convergence range. Under varying initial conditions, the system evolves into different dynamic regimes. The second method changes the connection topology of the system, which leads to stronger oscillatory behavior. In practical applications, the selection between these modes should be tailored to the specific functional demands of the system.

To better capture the system's evolving behavior, we introduced the concept of on-road energy, a novel global metric that estimates the system's developmental trajectory. Compared to analyzing individual state variables in time series, this measure offers a faster and more robust indication of the oscillatory status of the system.

Finally, we demonstrated that the introduction of an external input of sufficient strength can effectively suppress oscillations and stabilize the system at a new fixed point. This controllability highlights the model's strong representational capacity and its biological plausibility.

In a summary, the model's functional flexibility and tunability closely mirror the dual behavior observed in biological neurons—namely, the ability to both respond to external stimuli and autonomously generate rhythmic activity. This alignment provides a novel avenue for constructing biologically plausible neural models. Our findings not only extend our prior work on signal propagation but also lay a theoretical foundation for the future development of efficient, low-power, and structurally unified neuromorphic systems. These results are particularly promising for applications in neuromorphic computing, rhythmic signal generation, and brain–machine interfaces.

## Acknowledgements

Thanks to China Scholarship Council (CSC) for their support during the pandemic, which allowed me to get through those difficult days and give me the opportunity to put my past ideas into practice, ultimately resulting in the article I am sharing with you today.